# Stroke Prediction using Clinical and Social Features in Machine Learning

Aidan Chadha
Virginia Tech
Blacksburg, VA
aidanchadha03@vt.edu

## Abstract

*Every year in the United States, 800,000 individuals suffer a stroke - one person every 40 seconds, with a death occurring every four minutes. While individual factors vary, certain predictors are more prevalent in determining stroke risk. As strokes are the second leading cause of death and disability worldwide, predicting stroke likelihood based on lifestyle factors is crucial. Showing individuals their stroke risk could motivate lifestyle changes, and machine learning offers solutions to this prediction challenge. Neural networks excel at predicting outcomes based on training features like lifestyle factors, however, they're not the only option. Logistic regression models can also effectively compute the likelihood of binary outcomes based on independent variables, making them well-suited for stroke prediction. This analysis will compare both neural networks (dense and convolutional) and logistic regression models for stroke prediction, examining their pros, cons, and differences to develop the most effective predictor that minimizes false negatives.*

## 1. Introduction

According to the CDC, every year in the United States, 800,000 individuals suffer a stroke [1]. These strokes rely on a multitude of both external and internal factors. Some external factors include whether an individual has been married, what type of work they perform, where an individual chooses to live, etc. Some of the internal factors are whether the individual has heart disease, what the individuals BMI is, age, etc. These are all measurable factors that are known to increase an individual's likelihood for a stroke, and data sets have been collected with these measurables in mind. These data sets can then be used to assess whether individuals are more or less likely to have strokes based off the measurables.

Currently in the computational health care space, there is work ongoing to create more stroke prediction machine learning models. See the *Additional Readings* section for a few examples of these papers, as well as extra resources. These predictors, when integrated into clinical settings, could provide real-time risk assessments during routine check-ups. There have been predictors created using a perceptron, and dense and convolutional neural networks. This paper will allow for further insight into those results and their ability to avoid false negatives, which is crucial in healthcare settings where missing potential stroke risks could have life-threatening consequences.

There are a multitude of ways to be able to analyze these data sets to be able to create predictors, and for this project, there are two that will be focused on: logistic regression and neural networks. Within neural networks, there will be a focus on two architectures, convolutional and dense nets, to compare their differences in both accuracy and computational cost for training them. These three models will be trained using a Stroke Prediction Dataset collected from Kaggle aggregated by a data scientist at Kaggle. This data set will contain ~5000 individuals, each with their own stroke predictors, and with a binary classification of whether that individual had a stroke. If successful with a high enough accuracy, this will allow for healthcare professionals to be able to take in a survey of physical and social markers from an individual, and accurately predict if that individual will have a stroke, allowing them a chance to change their habits before it is too late.

The models will be trained to be able to accurately predict the binary (0, 1) result of whether the individual has or has not had a stroke. The data set used will be expanded on in the upcoming sections. For the models themselves, PyTorch as well as Sklearn will be used for the neural network and logistic regression models respectively.

All code is published to GitHub at the following link: https://github.com/Aidan7757/stroke_prediction_using_clinical_social_features.

All parameters are saved within the code, and the code is available for reproducibility with the model parameters being seed set, and the data files are also provided within the repository. The code is licensed by the MIT Open-Source License and was created from scratch.

For these models, the focus will be on reducing the number of false negatives within the predictions. This is due to the health care focus of the models, where false negatives are more detrimental to an individual's well-being, as it allows them to continue harmful day to day lifestyle choices, rather than pushing for a change that would decrease their chance of stroke.

1.1. Stroke Prediction Data Set

The stroke prediction data set was created four years ago and focuses on the internal and external factors that were discussed in the introduction. The exact factors are the numerical variables of the age of the individual, if the individual has hypertension (binary), if the individual has heart disease (binary), average glucose level, and their BMI. For the non-numerical columns: gender, if the individual has been married or not, the type of work the individual performs (Private, Public, Self-Employed), and the residence type of the individual (Rural, Urban). These different features will serve as the inputs into the models to be able to predict whether the individual will have a stroke, using the binary stroke variable, 0 for if the individual has not had a stroke, and 1 if they have. In practice, every non-numeric data point was encoded to be between 0 and the number of potential options for the variable – 1, and this was done using the Sklearn Preprocessing Label Encoder, so the models were able to accurately utilize these non-numeric variables. The data set was split into both a training and a testing data set using Sklearn, with testing data being 20% of the overall data set, and training being the remaining 80% through the Sklearn data split function.

One key issue with the data set is the difference in the amount of stroke cases vs. non-stroke cases. Only ~5-6% of the data set is of stroke cases, with the remaining being non-stroke cases. This is due to the rarer observations being of individuals who have had strokes and needs to be accounted for in analyzing the results of the model as the model will perform better at predicting whether an individual will not have a stroke, rather than having one.

Below are the plots of the different numeric non-binary variables within the data set.

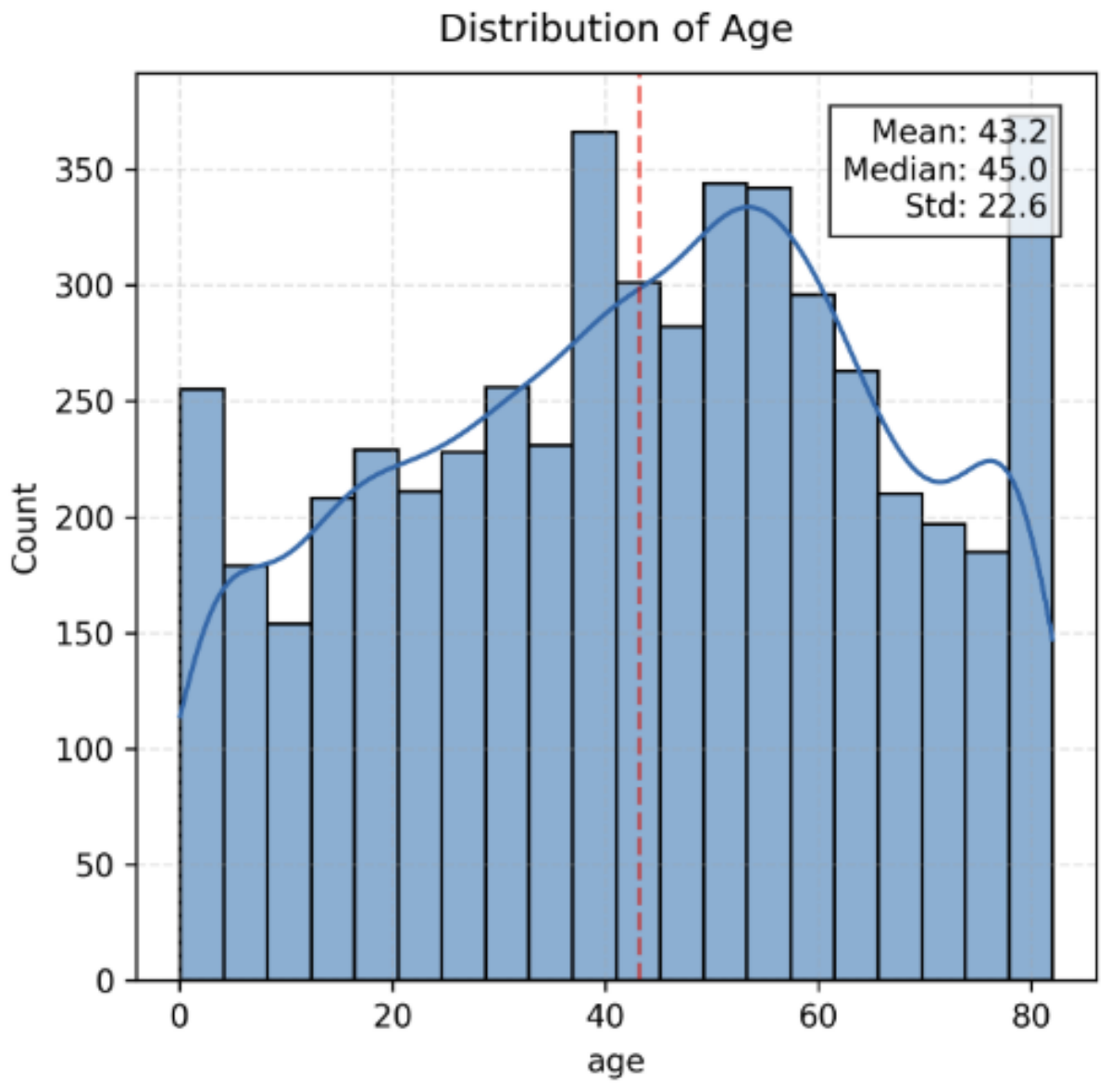

Figure 1: Distribution of the ages within the data set.

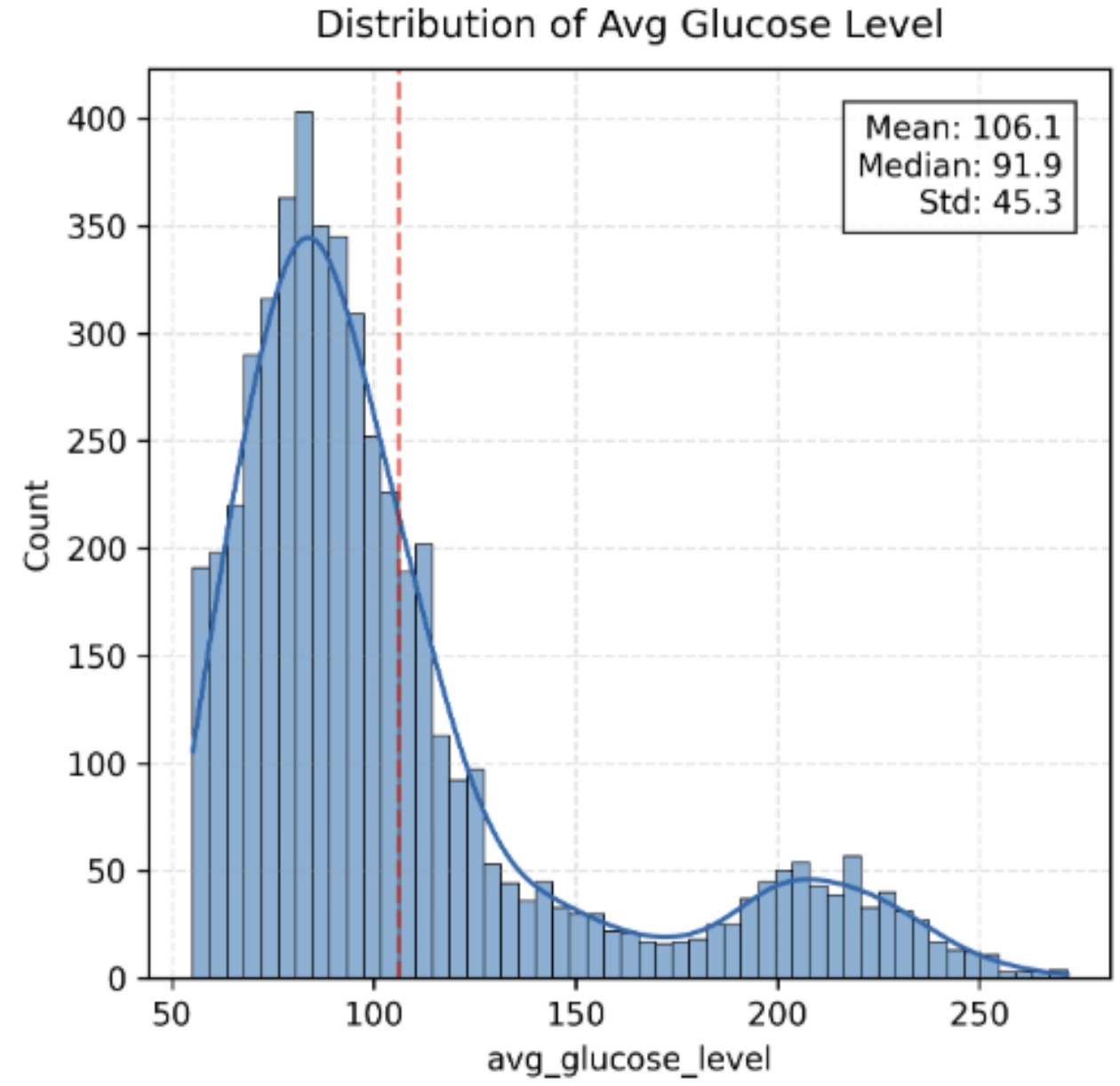

Figure 2: Distribution of average glucose level within the data set.

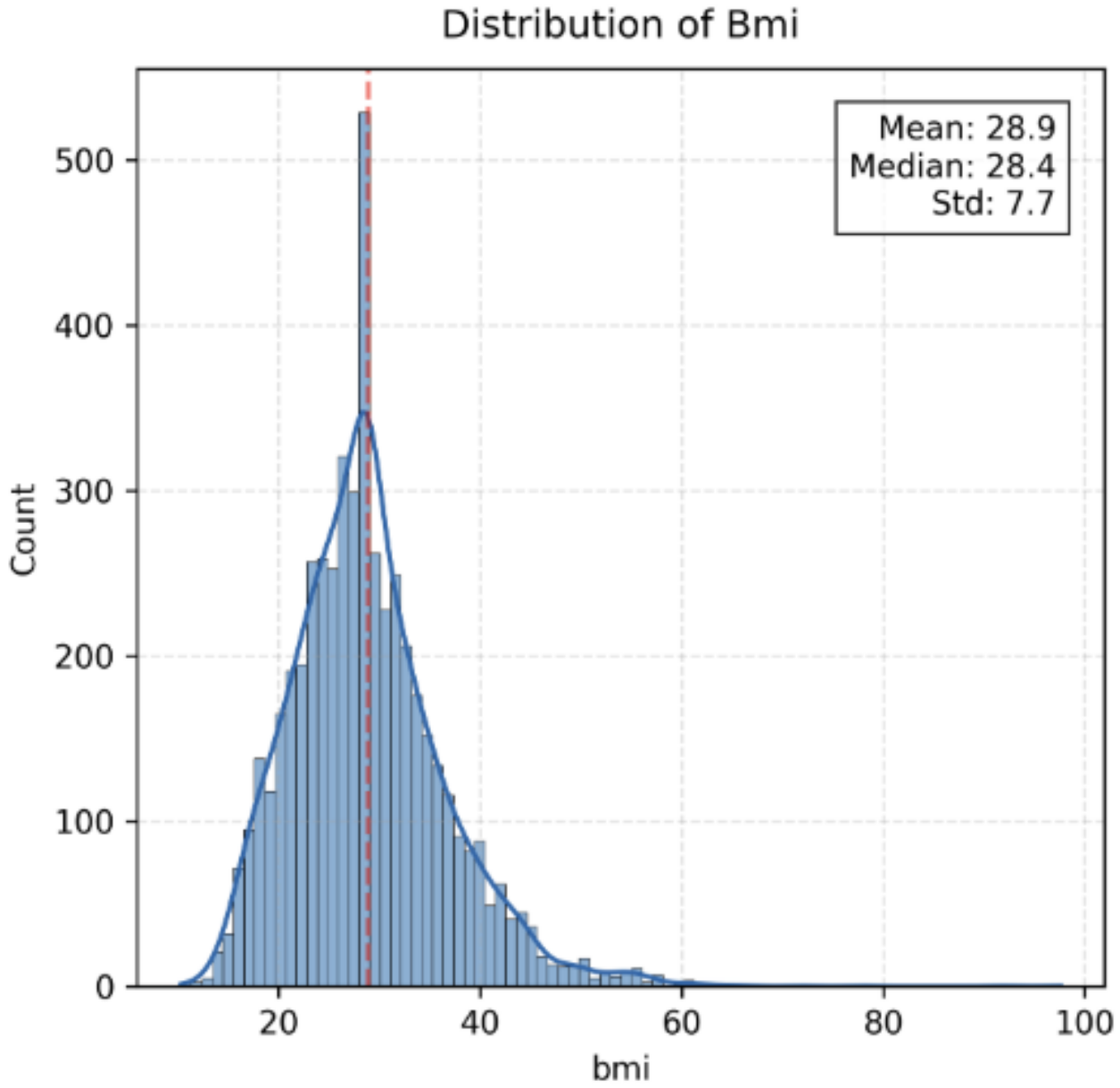

Figure 3: Distribution of the BMI within the data set.

From current stroke prediction research, biological factors are some of the most accurate predictors of whether an individual will have a stroke, namely BMI and age, which is stated by Shao et al. [2]. From this, it is predicted that the models will mainly use these numeric biological predictors as the best determinant of if an individual will have a stroke or not.

## 2. Logistic Regression Model Overview

Logistic regression is particularly well-suited for the stroke prediction task, where we process ten distinct medical features through standardized scaling using Sklearn's StandardScaler. For data preprocessing, all non-numeric variables were encoded using label encoding. The model applies weights to each standardized feature and combines them linearly before passing through a sigmoid function, which transforms the output into a probability between 0 and 1. The model employs L2 regularization (controlled through the default parameters in Sklearn's Logistic Regression) to prevent overfitting by penalizing large coefficients. The optimization process uses maximum iterations of 10,000 to ensure convergence, as lower iteration values failed to converge. The decision boundary is set at 0.5 probability, though this threshold can be adjusted to optimize for either sensitivity or specificity depending on clinical requirements.

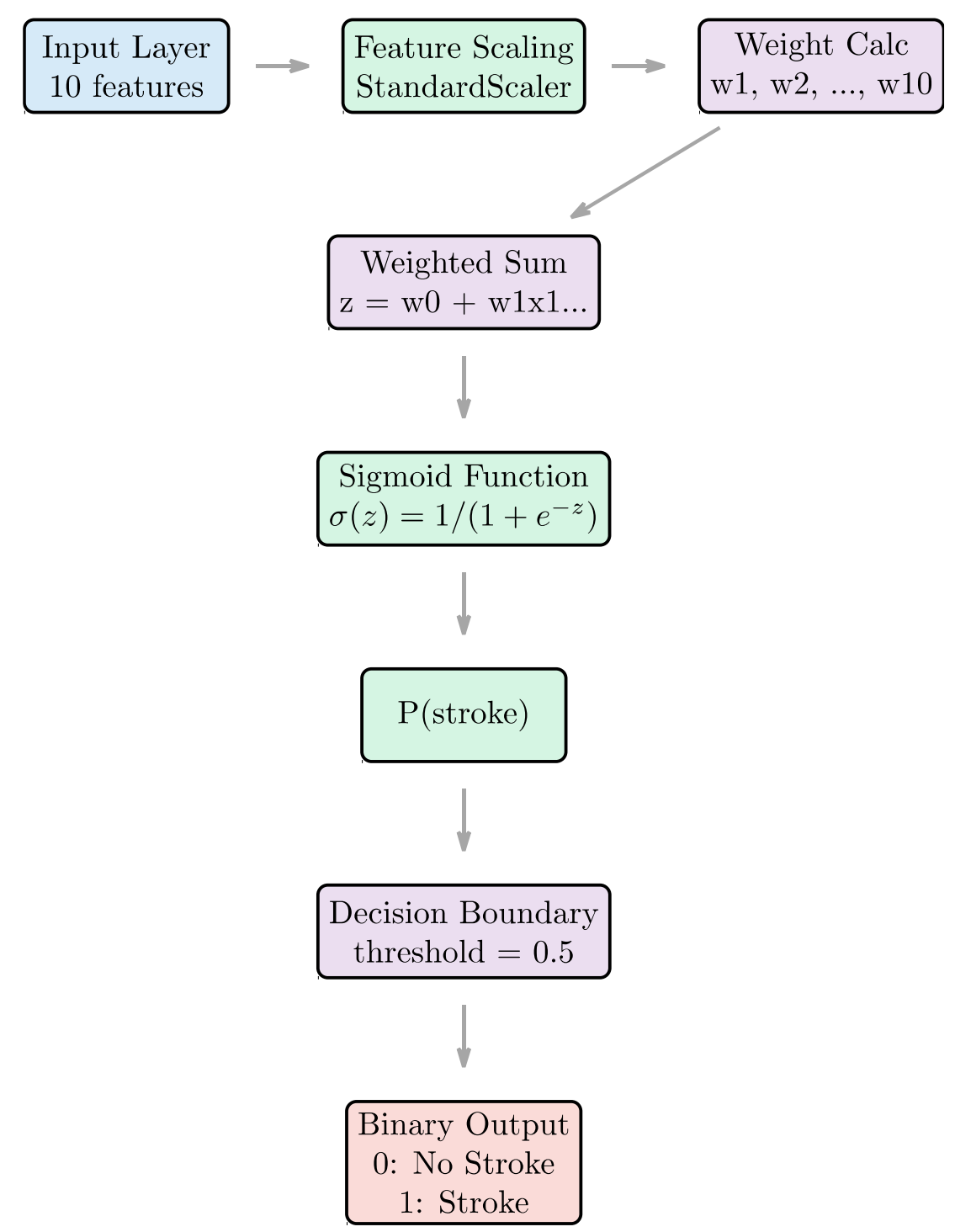

Figure 4: Graph of logistic regression model and each layer of the model, resulting in a binary output of the stroke prediction. Blue is the input stage, green is the processing stage, purple is the processing stage, and red is the output stage.

## 2.1. Logistic Regression Model Results

The logistic regression model employs multiple metrics to provide a comprehensive evaluation of stroke prediction performance. For the logistic regression model, accuracy, recall, precision, AUC, ROC, and the confusion matrix are all found using Sklearn's premade packages. Accuracy measures overall correctness but can be misleading with imbalanced data, which is why recall was included to assess the model's ability to identify actual stroke cases - critical in medical contexts where false negatives can have severe consequences. Precision helps balance this by measuring false positive rate, as unnecessary medical interventions also carry risks and costs. The AUC-ROC score evaluates model performance across different classification thresholds, providing insight into the model's discriminative ability regardless of the chosen probability cutoff. The confusion matrix offers detailed visualization of misclassifications, while feature coefficients reveal the relative importance of different medical factors, helping validate the model against clinical knowledge. Feature coefficients also allow us to gauge whether the initial research of BMI, age, and other biological factors being most critical to stroke prediction was correct.

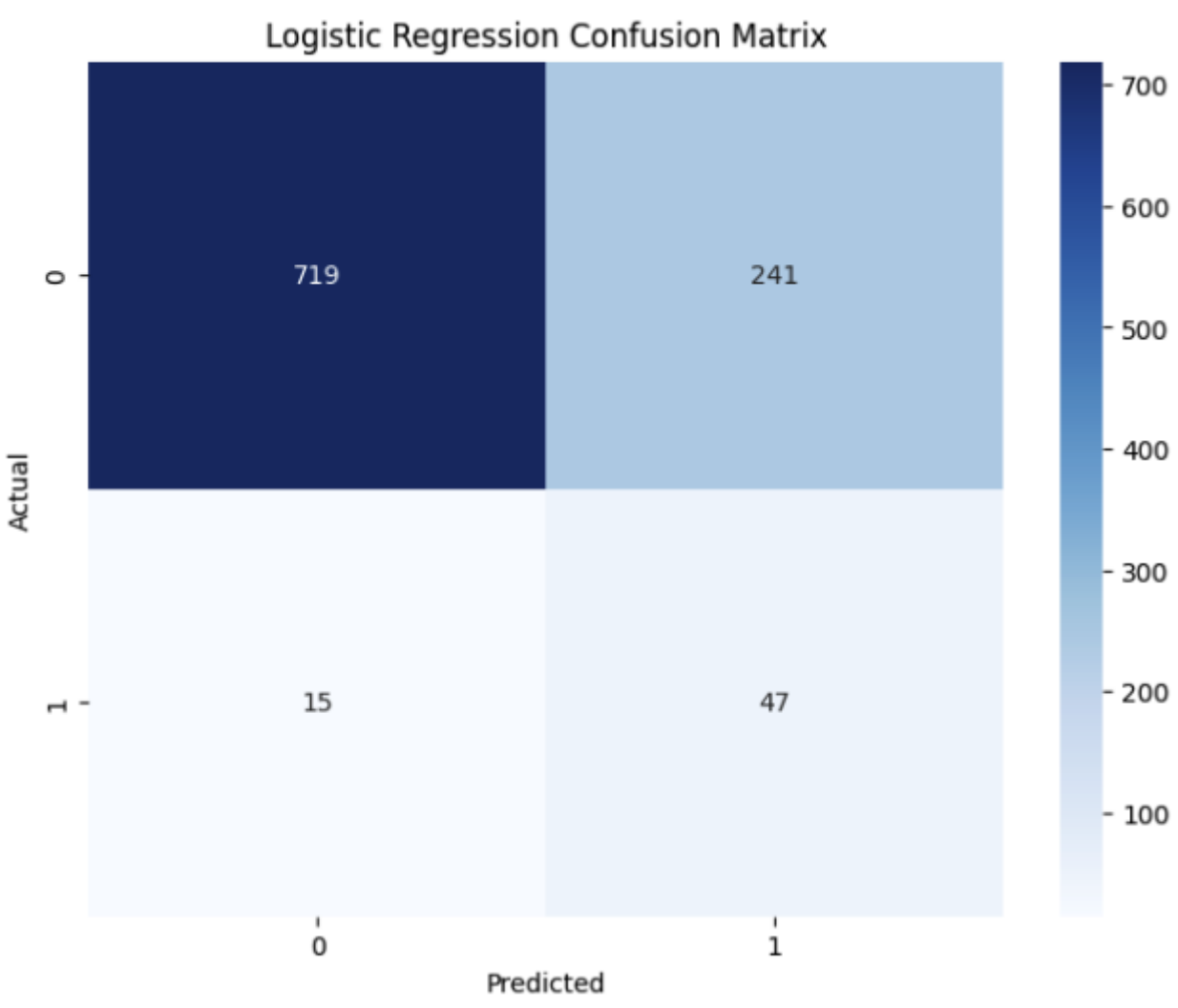

Figure 5: Confusion matrix of the logistic regression model. This shows that most incorrect predictions were arising from false positives. This prediction could be due to the imbalance in the data, as stroke cases with the data collected are rarer.

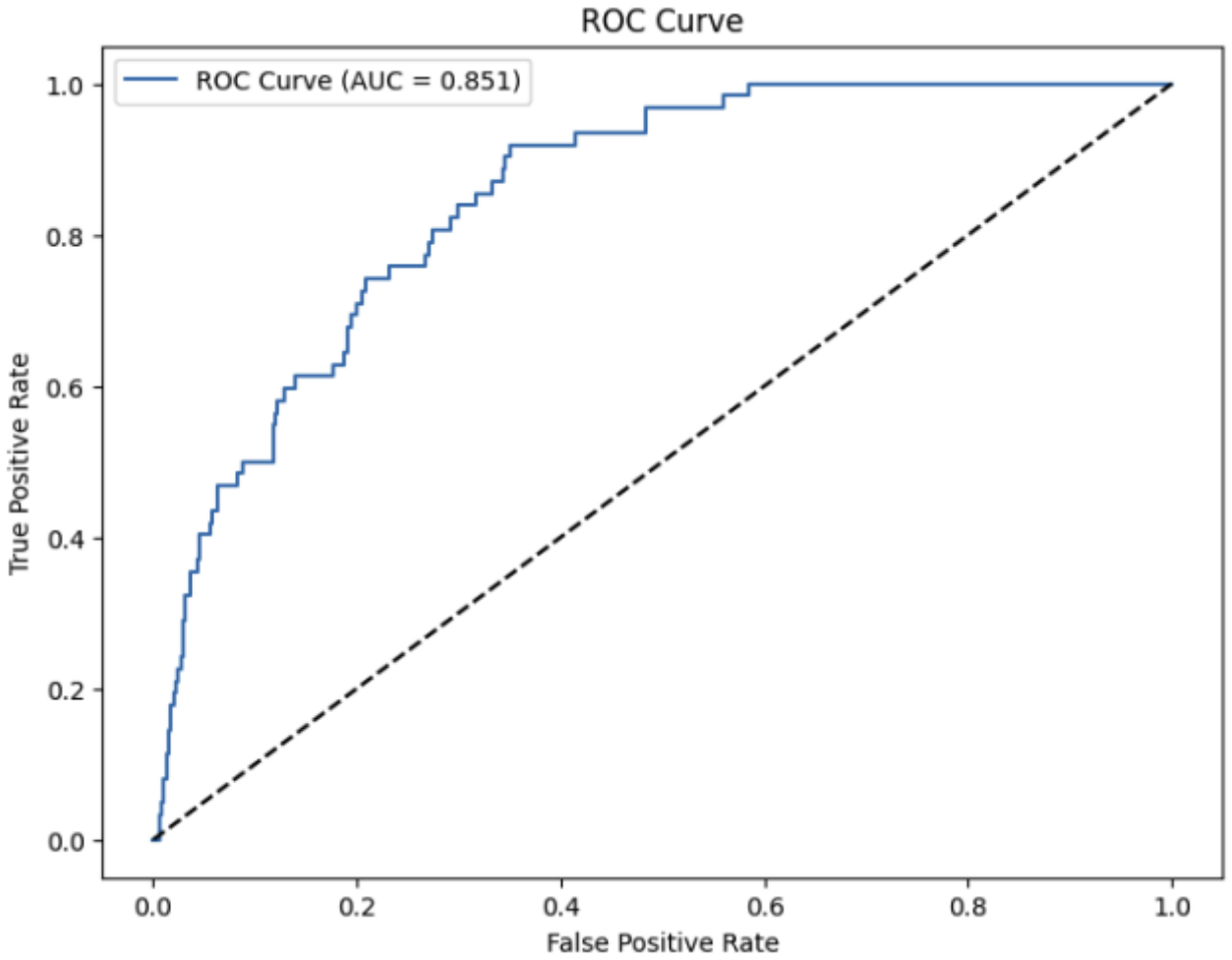

Figure 6: ROC Curve of the logistic regression model. The blue curve demonstrates that the logistic regression model is performing well at being able to detect true stroke cases at different decision thresholds.

From the feature coefficient absolute values, it is seen that true to the background knowledge, age is the top predictor of stroke. However, BMI is quite low on the feature importance list for the logistic regression model that is currently being trained, contrary to the background knowledge on stroke prediction. This means that the model is undervaluing what should be one of the top predictors of whether an individual will have a stroke and may contribute to a lower accuracy.

For the accuracy of the model, the model obtained a 74.95%. For recall, a result of 75.81% was obtained, and for precision, a result of 16.31%. One of the alarming results from this model is the precision, and this is due to a high false positive rate. Only 16.31% of predicted stroke cases are actual strokes, and this is believed to be due to the tradeoff of wanting to minimize the false negatives of the model. Overall, the model performs quite well in accuracy and recall, despite the imbalanced data set.

| Metric | Value (%) | Assessment |
|---|---|---|
| Accuracy | 74.95 | Strong baseline given imbalanced data |
| Recall | 75.81 | High stroke detection rate |
| Precision | 16.31 | Significant false positive rate |
| AUC-ROC | 85.06 | Good discriminative ability |

Table 1: Logistic Regression Model Performance Metrics

## 3. Neural Network Models Overview

When developing the neural networks, the decision was made to test dense and convolutional neural network models. These are two model most popular, as well as the theoretically simplest neural networks to develop and train. These models were developed using the Pytorch package. These neural networks are expected to perform better than the logistic regression model due to their complex architectures and tuned hyperparameters. Cross entropy was chosen as the loss function as it assigns a large penalty when the model is very confident about its prediction but wrong, helping reduce false negatives while naturally handling class imbalance through log-scaling. The models also implement dropout and batch normalization between layers to prevent overfitting and use the Adam optimizer to implement momentum within the gradient descent.

| Parameter | CNN | Dense NN |
|---|---|---|
| Input Size | 10 | 10 |
| Hidden Layers | 3 | 3 |
| Activation | ReLU | ReLU |
| Regularization | Dropout(0.3) | Dropout(0.3) + BatchNorm |
| Pooling | MaxPool1d(2) | None |
| **Training Parameters** | | |
| Learning Rate | 0.01 | |
| Epochs | 400 | |
| Batch Size | 32 | |
| Optimizer | Adam | |
| Loss Function | CrossEntropy with class weights | |

Table 2: Neural Network Architecture and Training Parameters

Next are two graphs of each neural network (DNN, CNN) and their respective architectures. The Dense Neural Network employs a narrowing architecture (128->64->32->2 neurons) to refine the features for classification. Each layer uses batch normalization and dropout (0.3) to prevent overfitting, with ReLU activation enabling non-linear transformations. The CNN architecture combines convolutional layers (16 and 32 filters) to extract spatial features from the input data, using max pooling to reduce dimensionality. It then flattens these features and passes them through fully connected layers (128->64->2) with dropout regularization. Both networks end with two output neurons for binary classification of stroke risk.

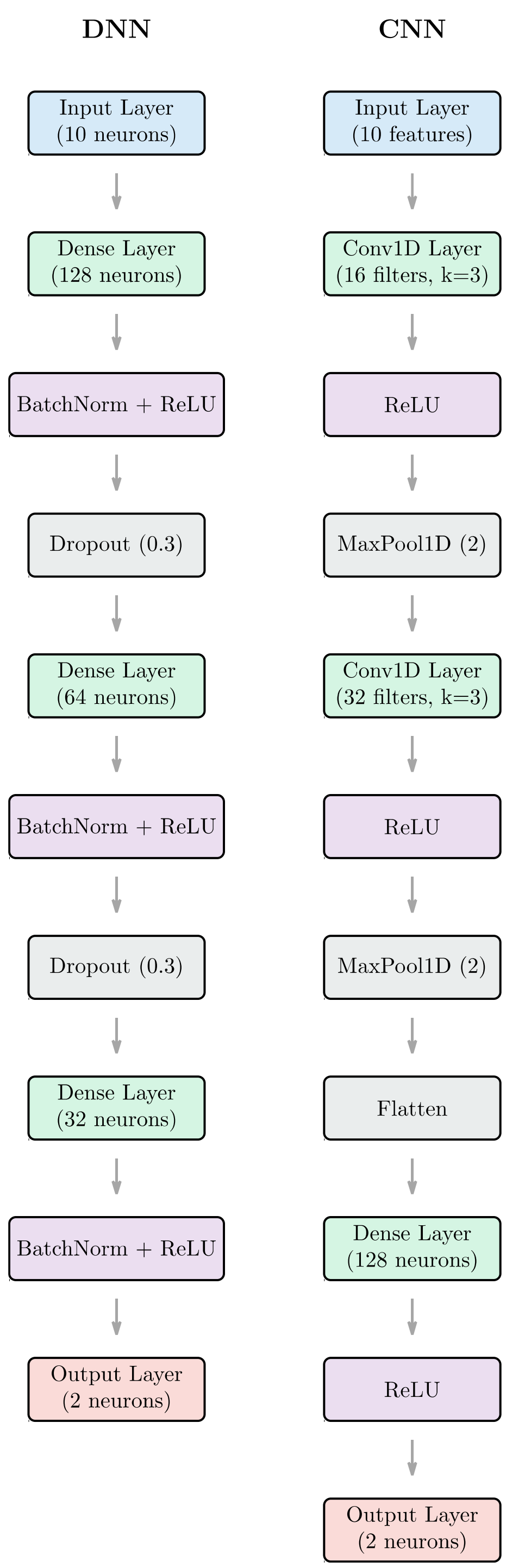

Figure 7: Layers of both the dense and convolutional neural networks. Batch Norm was included within the dense neural networks to allow for a higher learning rate to be chosen while keeping stability. Blue is the input layer, green is the hidden layers, purple is activation and normalization layers, gray is regularization layers, and red is output layers.

## 3.1. Neural Network Models Results

For neural network evaluation, multiple metrics were tracked across both training and testing phases. All metrics are tested using Sklearn’s prebuilt packages and can be seen within the GitHub repository. Training and testing accuracy measure general performance, while F1 scores provide a balanced measure of precision and recall, crucial for imbalanced datasets. The model's training loss helps monitor convergence and potential overfitting. The confusion matrix visualizes prediction errors, while precision and recall history track the model's ability to identify stroke cases over time. This comprehensive set of metrics allows comparison between epochs and between different neural network architectures (CNN vs Dense), while maintaining focus on the critical medical need to minimize false negatives in stroke prediction.

First is an overview of all the metrics for both the dense and convolutional neural networks as they progress through the 400 training epochs.

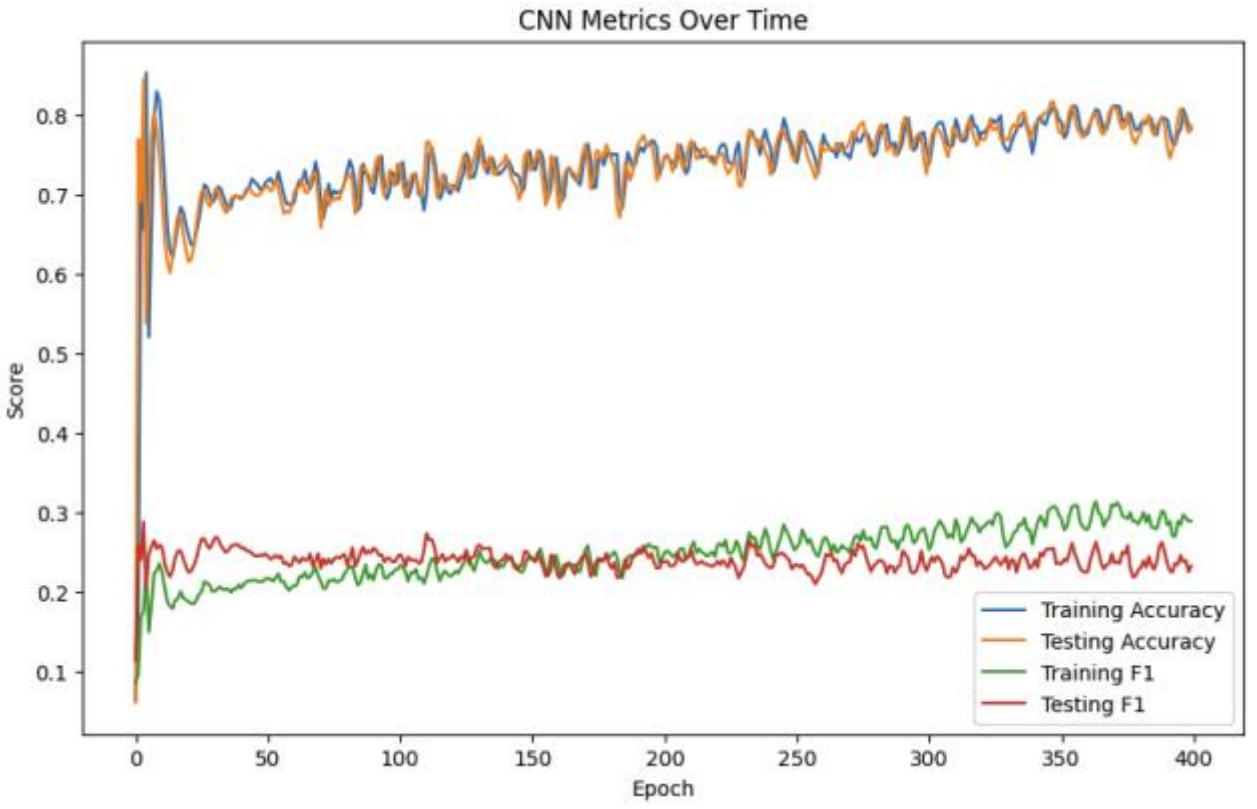

Figure 8: Accuracy and F1 for both testing and training data over time as the model progressed through the epochs. There is a significant slowdown after 50 epochs, signaling that 400 epochs may be inconsequential to the CNN network, but with the potential of leading to overfitting. Despite the potential overfitting, the model has very similar training and testing accuracy.

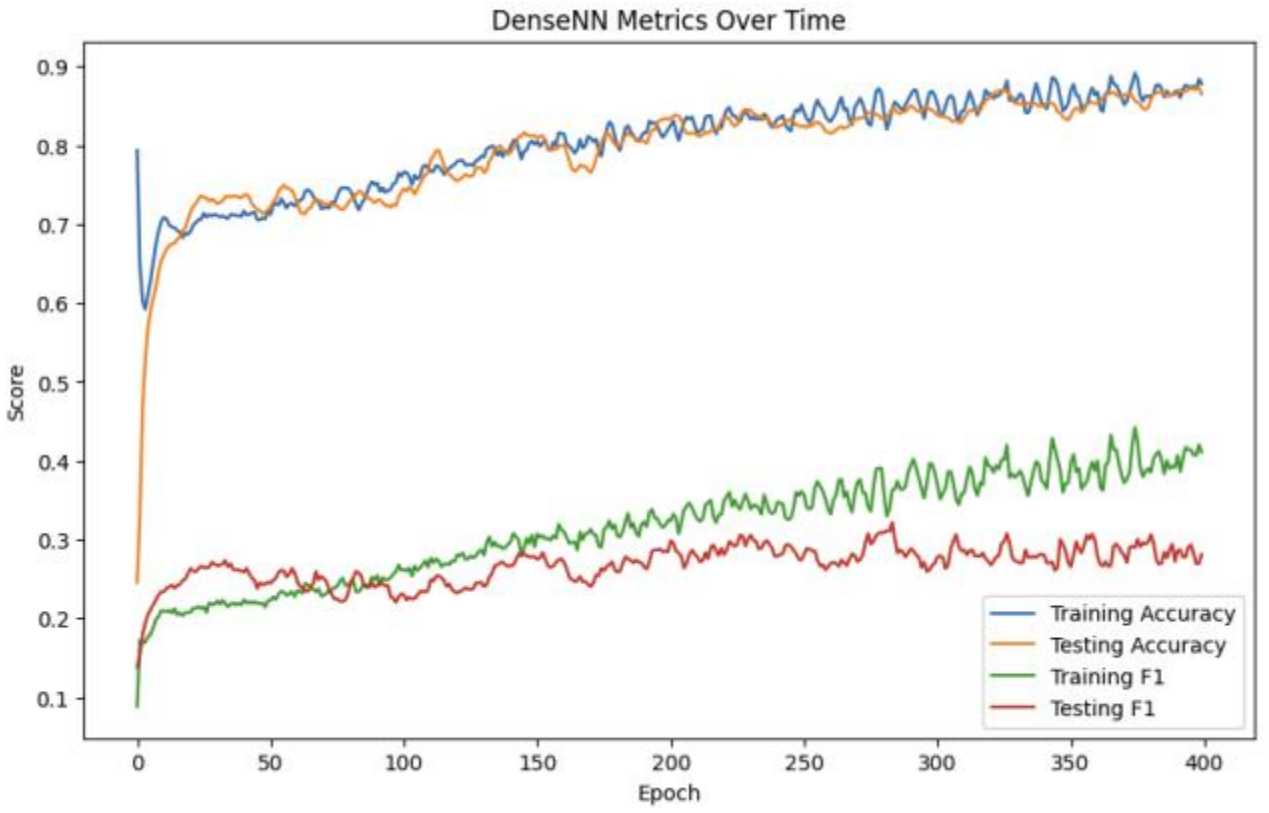

Figure 9: Accuracy and F1 for both testing and training data over time as the model progressed through the epochs. Unlike the CNN, the DNN model continues to increase in accuracy and F1 score as the model progresses through all 400 epochs, and this may be due to the Batch norm.

Despite lower than desired accuracy, there is still interest in being able to compare the performance of the DNN vs the CNN in the different measurables.

| **Metric** | **CNN** | **Dense NN** |
|---|---|---|
| Accuracy (%) | 78.67 | 86.50 |
| Precision (%) | 14.86 | 20.77 |
| Recall (%) | 53.23 | 43.55 |
| F1 Score | 0.2324 | 0.2812 |
| Training Time (s) | 37.77 | 16.77 |

Table 3: Performance Comparison of Neural Network Models

Overall, the dense neural network seems to perform better in nearly every metric, except for a worse result in recall. Recall means that the dense neural network is performing worse in being able to predict more true stroke cases. This is an important factor in picking which model to use, as while dense neural networks perform better overall, failing to predict true stroke cases contributes to a lack of true relevancy of the model, as it fails on the metric where the model would be the most helpful within a healthcare setting.

From the analysis of the accuracy over the epochs in the model training, there does not seem to be overfitting within either the dense or convolutional models, as the testing and training data accuracies grew together, so any overfitting was inconsequential. As the number of epochs grew, both the convolutional and dense neural networks accuracies grew.

When looking at the time for the models to each train, the dense neural network trains quicker than the convolutional neural network. This is believed to be due to the simpler nature of the layers of the dense neural network, despite containing a higher accuracy and overall better performance than the convolutional neural network.

## 4. Conclusion

In comparing the three models for stroke prediction (Logistic Regression, Dense Neural Net, Convolutional Neural Net), each offers distinct advantages depending on the metrics where it performs best. To improve the accuracy of these models, what is believed to be the most important next step would be more data regarding true patients who have had strokes, and their life habits leading up to it. This will help some of the imbalance in the data that is currently present and will allow the models to more accurately predict the stroke likelihood. Logistic regression provides easily interpretable results with reasonable accuracy (74.95%) and high recall (75.81%), making it valuable for initial risk screening. The Dense Neural Network achieved the highest accuracy (86.50%) and precision (20.77%), suggesting its use may be best for detailed patient assessment where false positives are more severe. The CNN, with its accuracy (78.67%) but higher recall (53.23%) than the Dense NN, could serve as a more comprehensive approach.

Healthcare implementations may benefit from a multi machine learning model system: using logistic regression for initial screening due to its interpretability and high recall, followed by the Dense Neural Network for high-risk cases requiring more precise assessment, with the CNN serving as a validation tool given its balanced performance profile, creating a comprehensive stroke risk assessment pipeline that leverages each model's strengths while mitigating their individual weaknesses.

## 6. Additional Readings